%% file: main.tex
\documentclass[10pt,twocolumn,letterpaper]{article}

\usepackage[pagenumbers]{cvpr} 

\usepackage{indentfirst} 
\newcommand{\method}{\texttt{VideoLCM}\xspace}

\input{preamble}

\definecolor{cvprblue}{rgb}{0.21,0.49,0.74}
\usepackage[pagebackref,breaklinks,colorlinks,citecolor=cvprblue]{hyperref}
\usepackage{capt-of}
\usepackage[dvipsnames]{xcolor}
\usepackage{xspace}
\usepackage{enumitem}
\usepackage{amsmath,amssymb,amsbsy,amsfonts,dsfont,pifont,bm,bbm,mathrsfs,mathtools,nicefrac}
\usepackage{algorithm,algpseudocode,listings}
\usepackage{booktabs,multirow,adjustbox,diagbox,threeparttable,tabularray,colortbl}

\definecolor{Gray}{gray}{0.94}
\definecolor{liGray}{gray}{0.5}
\definecolor{LightCyan}{rgb}{0.88,1,1}
\newcommand{\tablestyle}[2]{\setlength{\tabcolsep}{#1}\renewcommand{\arraystretch}{#2}\centering\small}

\newlength\savewidth\newcommand\shline{\noalign{\global\savewidth\arrayrulewidth
  \global\arrayrulewidth 1pt}\hline\noalign{\global\arrayrulewidth\savewidth}}
  
\def\paperID{*****} 
\def\confName{CVPR}
\def\confYear{2024}

\title{VideoLCM: Video Latent Consistency Model}

\author{
   Xiang Wang$^{1*}$
     \hspace{0.01cm} 
    Shiwei Zhang$^{2\dag}$
     \hspace{0.01cm} 
    Han Zhang$^{3}$
     \hspace{0.01cm} 
    Yu Liu$^2$ 
     \hspace{0.01cm}
    Yingya Zhang$^2$ 
     \hspace{0.01cm}
    Changxin Gao$^1$ 
     \hspace{0.01cm}
     Nong Sang$^{1\dag}$
      \vspace{2mm}
     \\
    $^1$Key Laboratory of Image Processing and Intelligent Control,\\   School of Artificial Intelligence and Automation, Huazhong University of Science and Technology\\
      $^2$Alibaba Group \hspace{0.5cm} $^3$Shanghai Jiao Tong University \\
{\tt\scriptsize \{wxiang,cgao,nsang\}@hust.edu.cn, \{zhangjin.zsw,ly103369,yingya.zyy\}@alibaba-inc.com, hzhang9617@gmail.com}
}

\begin{document}
%

\twocolumn[{
\maketitle
    \centering
    \vspace{-12pt}
    \includegraphics[width=0.95\textwidth]{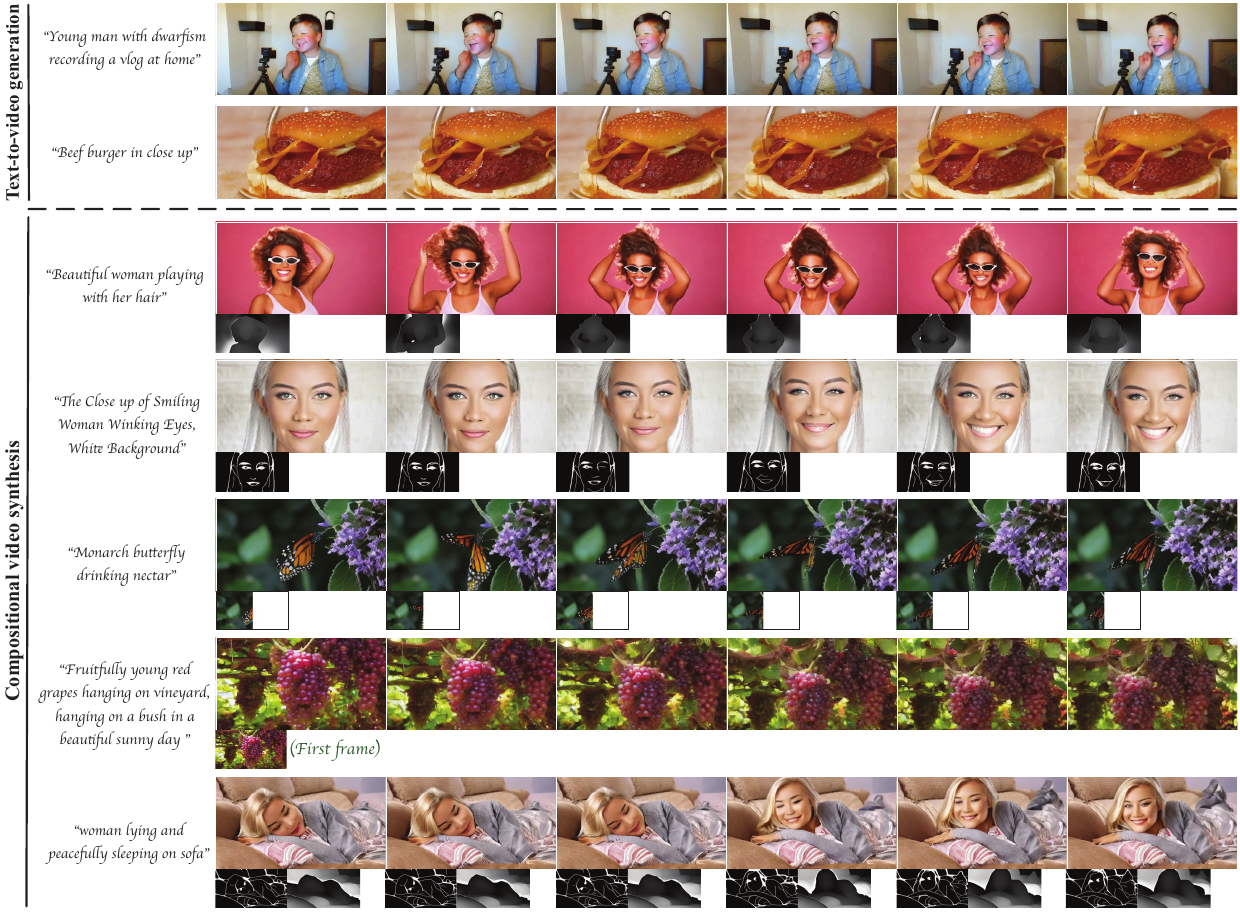}
    \vspace{-5pt}
    \captionof{figure}{
        \textbf{Video examples} synthesized by the proposed \method with \emph{4 inference steps}.
        \method is a plug-and-play technique and can be integrated into text-to-video generation and compositional video synthesis paradigms.
        %
        %
    }
    \label{first_figure}
    \vspace{15pt}
    }
]

\let\thefootnote\relax\footnotetext{$*$ Intern at Alibaba Group. \hspace{1mm} $\dag$ Corresponding authors. }
\input{sec/0_abstract}    
\input{sec/1_intro}

\input{sec/2_related_work}

\input{sec/3_method}

\input{sec/4_experment}
\input{sec/5_conclusion}
{
    \small
    \bibliographystyle{ieeenat_fullname}
    \bibliography{main}
}


\end{document}

%% file: preamble.tex
\usepackage[dvipsnames]{xcolor}


%% file: sec/0_abstract.tex
\vspace{-10mm}
\begin{abstract}

Consistency models have demonstrated powerful capability in
efficient image generation and allowed synthesis within a few sampling steps, alleviating the high computational cost in diffusion models.
However, the consistency model in the more challenging and resource-consuming video generation is still less explored.
In this report, we present the \method framework to fill this gap, which leverages the concept of consistency models from image generation to efficiently synthesize videos with minimal steps while maintaining high quality. 
\method builds upon existing latent video diffusion models and incorporates consistency distillation techniques for training the latent consistency model.
%
Experimental results reveal the effectiveness of our \method in terms of computational efficiency, fidelity and temporal consistency.
%
Notably, \method achieves high-fidelity and smooth video synthesis with only four sampling steps, showcasing the potential for real-time synthesis.
We hope that \method can serve as a simple yet effective baseline for subsequent research.
The source code and models will be publicly available.

\end{abstract}

%% file: sec/1_intro.tex
\section{Introduction}
\label{sec:intro}

Recently, the field of video generation has witnessed tremendous advancements in synthesizing photo-realistic and temporally coherent video content, especially with the development of diffusion models~\cite{videocomposer,modelscopet2v,yin2023dragnuwa,chai2023stablevideo,make-a-video,guo2023animatediff,SVD,hu2023animate,zhang2023i2vgen,TFT2V,zhao2023controlvideo}.
Traditional diffusion-based methods such as videoLDM~\cite{VideoLDM}, Make-A-Video~\cite{make-a-video} and ModelScopeT2V~\cite{make-a-video}, have achieved significant performance by incorporating additional temporal layers into existing image diffusion models~\cite{stablediffusion,Dalle2} to handle the temporal dynamics in videos. 
%
Nevertheless, these diffusion-based approaches inevitably require substantial sampling steps to synthesize videos during inference, \eg, 50-step DDIM sampling~\cite{DDIM}.
This limitation hinders the efficient and rapid synthesis of high-quality videos.

To address the challenge of high sampling cost in diffusion models, the concept of consistency models has been introduced in image generation~\cite{consistencymodel,LCM,LCM_lora,xiao2023ccm}, achieving remarkable progress by enabling efficient image synthesis with a minimal number of steps (\eg, 4 steps \vs 50 steps). 
%
%
Despite its success, the application of the consistency model in the domain of video synthesis still remains largely unexplored.

To fill this research gap, we propose the \method framework.
Our method builds upon existing latent diffusion models in video generation and leverages the idea of consistency distillation to train a video latent consistency model. 
By incorporating the \method framework, we aim to alleviate the need for extensive sampling steps while maintaining high-quality video synthesis. 
The quantitative and qualitative results demonstrate the effectiveness of our approach. 
Remarkably, our method achieves high-fidelity video synthesis with only 4$\sim$6 sampling steps, showcasing its potential for fast and real-time synthesis. 
In comparison, previous methods such as ModelScopeT2V~\cite{modelscopet2v} and VideoLDM~\cite{VideoLDM} typically require approximately 50 steps based on the DDIM solver to achieve similarly satisfactory results.
In addition to text-to-video generation, we further extend the consistency model to compositional video synthesis. 
Experimental results indicate that in compositional video synthesis tasks, such as compositional depth-to-video synthesis, \method can produce visually satisfactory and temporally continuous videos with even fewer steps, such as just 1 step.

In summary, the proposed \method bridges the gap between diffusion models and consistency models in video generation, enabling efficient synthesis of high-quality videos.
%
By exploring the potential of consistency models in video generation, we aim to contribute to the field of fast video synthesis and provide a simplified and effective baseline for future research.

%


%% file: sec/2_related_work.tex
\section{Related Work}

The relevant fields related to this work include text-to-image generation, consistency model, and video generation. Next, we provide a brief review of these fields.

\vspace{1mm}
\noindent \textbf{Text-to-image generation.} 
In recent years, significant progress has been made in image generation with the development of generative models~\cite{ho2022classifierfreeguidance,kang2023scaling,goodfellow2014generative,shen2021closed,chang2023muse,vprediction,chen2023anydoor,liu2023cones}, especially with the emergence of diffusion models~\cite{stablediffusion,saharia2022photorealistic,T2i-adapter,GLIDE,ruiz2023dreambooth,kawar2023imagic}. Typical methods for text-to-image generation, such as DALLE-2~\cite{Dalle2}, propose a two-stage approach where the input text is first converted into image embeddings using a prior model, followed by the generation of images based on these embeddings. 
Stable Diffusion~\cite{stablediffusion} introduces a VAE-based approach in the latent space to decrease computational demand and optimizes the model with large-scale datasets~\cite{schuhmann2022laion}.
Subsequent methods like ControlNet~\cite{controlnet} and Composer~\cite{huang2023composer} have incorporated additional conditional inputs, such as depth maps or sketches, for spatially controllable image synthesis. 
%


\vspace{1mm}
\noindent \textbf{Consistency model.} 
To alleviate the limitation of requiring a large number of inference steps in diffusion models, the consistency model~\cite{consistencymodel} has been developed. 
Built upon the probability flow ordinary differential equation (PF-ODE) , consistency models learn to map any point at any time step to the starting of the trajectory, \ie, the original clean image. 
The consistency model facilitates efficient one-step image generation without sacrificing the advantages of multi-step iterative sampling, thereby enabling more high-quality results through multi-step inference.
Building upon the foundation of the consistency model, LCM~\cite{LCM} explores consistency models in the latent space to save memory consumption and improve inference efficiency. Subsequently, several methods~\cite{LCM_lora,sauer2023adversarial,xiao2023ccm} have also investigated efficient generation techniques and achieved impressive results. 
Inspired by the efficiency of the consistency model and its significant success in image generation, we extend the application of the consistency model to the domain of video generation.

\begin{figure*}[t]
  \centering
  \includegraphics[width=0.90\linewidth]{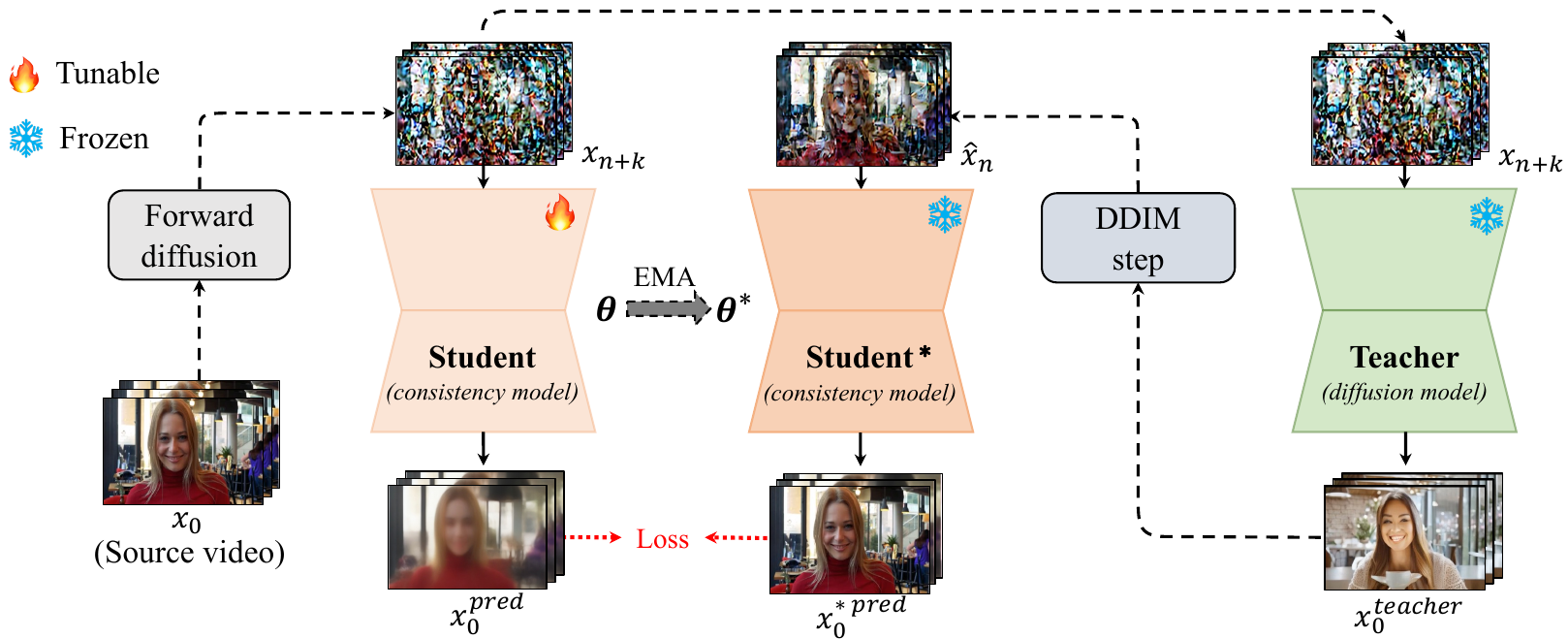}
  \vspace{-2mm}
    \caption{\textbf{The overall pipeline} of the proposed \method. Given a source video $x_{0}$, a forward diffusion operation is first performed to add noise to the video. Then, the noised $x_{n+k}$ is entered into the student and teacher model to predict videos. 
    $\hat{x}_{n}$ is estimated by the teacher model and fed into the EMA student model.
    To learn self-consistency, a loss is imposed to constrain the output of the two student models to be consistent. 
    The whole consistency distillation is conducted in the latent space, and conditional guidance is omitted for ease of presentation.  The teacher model is a video diffusion model, and the student model shares the same network structure as the teacher model and is initialized with the parameters of the teacher model.
    }
    \label{fig:network}
\end{figure*}

\vspace{1mm}
\noindent \textbf{Video generation.} 
The pursuit of visually appealing and temporally coherent synthesis is crucial in video generation. Early methods~\cite{mocogan,wang2020g3an,balaji2019conditional,wang2020imaginator,wang2023styleinv,skorokhodov2022stylegan} primarily relied on generative adversarial networks (GANs), resulting in poor video quality and limited generalization to unseen domains~\cite{VideoLDM}. 
With the rise of diffusion models~\cite{stablediffusion}, which provide stable training and impressive results, recent approaches~\cite{modelscopet2v,tune-a-video,make-a-video,imagenvideo,Text2video-zero,ge2023preserve,molad2023dreamix,yu2023video,luo2023videofusion,chai2023stablevideo,guo2023animatediff,wang2023videofactory,SVD,an2023latent,zhou2022magicvideo,zhang2023show,zhang2023i2vgen,wei2023dreamvideo} have started exploring diffusion-based video generation. 
To generate temporally consistent videos, mainstream text-to-video methods such as ModelScopeT2V~\cite{modelscopet2v} and VideoLDM~\cite{VideoLDM} achieve long-term temporal modeling by inserting temporal modules into a 2D UNet and training on large-scale datasets~\cite{webvid10m}.
%
There are also some methods~\cite{make-a-video,molad2023dreamix,zhou2022magicvideo,villegas2022phenaki,zhang2023i2vgen} focusing on using super-resolution and frame interpolation models to generate more realistic videos. 
%
To achieve spatiotemporal controllability, controllable video synthesis methods~\cite{videocomposer,Gen-1,zhao2023controlvideo,zhang2023controlvideo,chen2023control,chen2023motion,ni2023conditional} have been proposed. In particular, VideoComposer~\cite{videocomposer} presents a compositional video synthesis paradigm that enables flexible control over the generated videos through textual, spatial, and temporal conditions.
%
Despite significant advancements, these methods rely heavily on extensive iterative denoising processes to obtain satisfactory results, posing challenges for fast video generation. 
To address this issue, in this work, we propose the \method framework based on consistency models for fast and efficient video synthesis.

%% file: sec/3_method.tex
\section{Method}

The proposed \method builds upon the foundations of latent consistency models. 
We first briefly describe the preliminaries about latent consistency models.
Then, we will present the details of the proposed \method. The overall structure of \method is displayed in \cref{fig:network}.

\subsection{Preliminaries}
To achieve fast image generation, song \etal~\cite{consistencymodel} brings into the conception of the consistency model, which aims to optimize a model that learns to map any point at any time step to the starting of the PF-ODE trajectory.
Formally, the self-consistency property can be formulated as:
\begin{equation}
    {{f_{\theta}}(x_{t},t) = {f_{\theta}}({x_{t^\prime}}, {t^\prime}), \forall t, {t^\prime} \in [\epsilon, T]}
    \label{Eq1}
\end{equation}
where $\epsilon$ is a time step, $T$ is the overall denoising step, and $x_{t}$ denotes the noised input.

To accelerate the training and extract the strong prior knowledge of the established diffusion model~\cite{stablediffusion}, consistency distillation is usually adopted:
\begin{equation}
    \mathcal{L} (\theta ,\theta^{*};\Phi )=\mathbb{E} [d({f_{\theta}}(x_{t_{n+1}}, {t_{n+1}})), {f_{\theta^*}}(\hat{x}_{t_{n}}, {t_{n}}))]
    \label{Eq2}
\end{equation}
where $\Phi$ means the applied ODE solver and the model parameter $\theta^*$ is obtained from the exponential moving average (EMA) of $\theta$.
$\hat{x}_{t_{n}}$ is the estimation of ${x}_{t_{n}}$:
\begin{equation}
    \hat{x}_{t_{n}}  \gets x_{t_{n+1}} + (t_{n} - t_{n-1})\Phi ({x_{t_{n+1}}} ,  {t_{n+1}})
    \label{Eq3}
\end{equation}
LCM~\cite{LCM} conducts the above consistency optimization in the latent space and applies classifier-free guidance~\cite{ho2022classifierfreeguidance} in \cref{Eq3} to inject control signals, such as textual prompts.
For more details, please refer to the original works~\cite{consistencymodel,LCM}.

\subsection{VideoLCM}

Following LCM, the proposed \method is also established in the latent space to reduce the computational burden.
%
%
To leverage the powerful knowledge within large-scale pretrained video diffusion models and speed up the training process, we apply the consistency distillation strategy to optimize our model.
Note that the pretrained video diffusion model can be the text-to-video generation model (\eg, ModelScopeT2V~\cite{modelscopet2v}) or the compositional video synthesis model (\eg, VideoComposer~\cite{videocomposer}).

In \method, we apply DDIM~\cite{DDIM} as the basic ODE solver $\Psi$ to estimate $\hat{x}_{t_{n}}$:
\begin{equation}
    \hat{x}_{t_{n}} \approx   x_{t_{n+1}} + \Psi (x_{t_{n+1}}, {t_{n+1}}, {t_{n}},c)
    \label{Eq4}
\end{equation}
where $c$ means the conditional inputs, which can be textual prompts in text-to-video generation or multiple combined signals in compositional video synthesis task.
%

Since classifier-free guidance is pivotal in synthesizing high-quality content~\cite{ho2022classifierfreeguidance}, we also leverage classifier-free guidance in the consistency distillation stage and use a factor $w$ to control the guidance scale:
\begin{equation}
\begin{split}
    \hat{x}_{t_{n}} \approx   & x_{t_{n+1}}  + (1+w) \Psi (x_{t_{n+1}}, {t_{n+1}}, {t_{n}},c) \\ & - w \Psi (x_{t_{n+1}}, {t_{n+1}}, {t_{n}},\phi)
\end{split}
    \label{Eq5}
\end{equation}
In LCM~\cite{LCM},  $w$ is variable and can be fed into the network for modulation, but this changes the structure of the initial network because a module encoding $w$ needs to be added. In order to keep the initial parameters and design of the consistency model consistent with the teacher diffusion model, we train the consistency model with a fixed value of $w$, such as $9.0$.
Note that classifier-free guidance is only applied to the teacher diffusion model in training and is not required during the inference process of the consistency model.

\method is a plug-and-play technique compatible with text-to-video generation and compositional video synthesis.
During the inference phrase, we can sample 4$\sim$6 LCM steps to produce plausible results on text-to-video generation.
For compositional video synthesis, take the compositional depth-to-video task as an example, 2$\sim$4 steps are sufficient, and sometimes even 1 step.
%

%% file: sec/4_experment.tex
\section{Experiments}

In this section, we first introduce the details of the experimental setup.
Then, quantitative and qualitative comparisons will be represented to evaluate the effectiveness of the proposed \method framework.

\subsection{Experimental setup}

\noindent \textbf{Datasets.}
We train our video consistency model on two widely used datasets, \ie, WebVid10M~\cite{webvid10m} and LAION-5B~\cite{schuhmann2022laion}. 
%
WebVid10M is a video-text dataset containing approximately 10.7M videos.
%
To further enhance temporal diversity and improve visual quality, we additionally utilize about 1M internal video-text data to train \method.
LAION-5B is an image-text dataset that is used to provide high-quality visual-text correspondence.

\noindent \textbf{Implementation details.}
%
%
For convenience, we directly leverage the existing pretrained video diffusion models in TF-T2V~\cite{TFT2V} as the teacher model and fix the network parameters in the consistency distillation process. 
TF-T2V~\cite{TFT2V} is a technique that exploits text-free videos to scale up video diffusion models and can be applied to mainstream text-to-video generation and compositional video synthesis framework.
The model structure of the consistency model is the same as the teacher diffusion model and is initialized with the teacher's model parameters.
AdamW optimizer with a learning rate of 1e-5 is adopted to train \method. 
The EMA rate used in our experiments is $0.95$.
We sample 16 frames and crop a $448 \times 256$ center region from each source video as model input.
The training loss used in \method is a Huber loss by default. 
%
The entire network is trained on 4 NVIDIA A100 GPUs (one for image and three for video), requiring approximately 4k training iterations to produce relatively reliable video generation results.

\begin{figure*}
  \centering
  \includegraphics[width=0.99\linewidth]{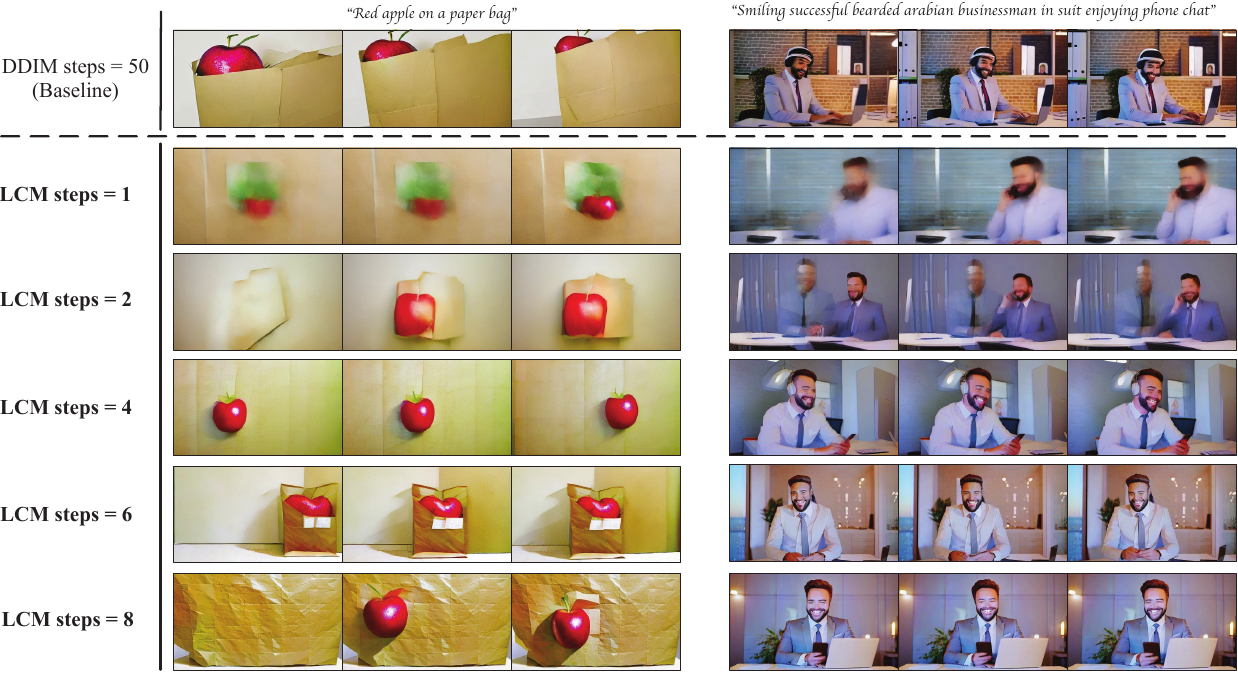}
  \vspace{-2mm}
    \caption{\textbf{Ablation study on text-to-video task under varying steps}. 
    Larger steps generally yield better visual quality and time continuity.
    }
    \label{fig:ablation_steps_t2v}
\end{figure*}

\begin{figure*}
  \centering
  \includegraphics[width=0.99\linewidth]{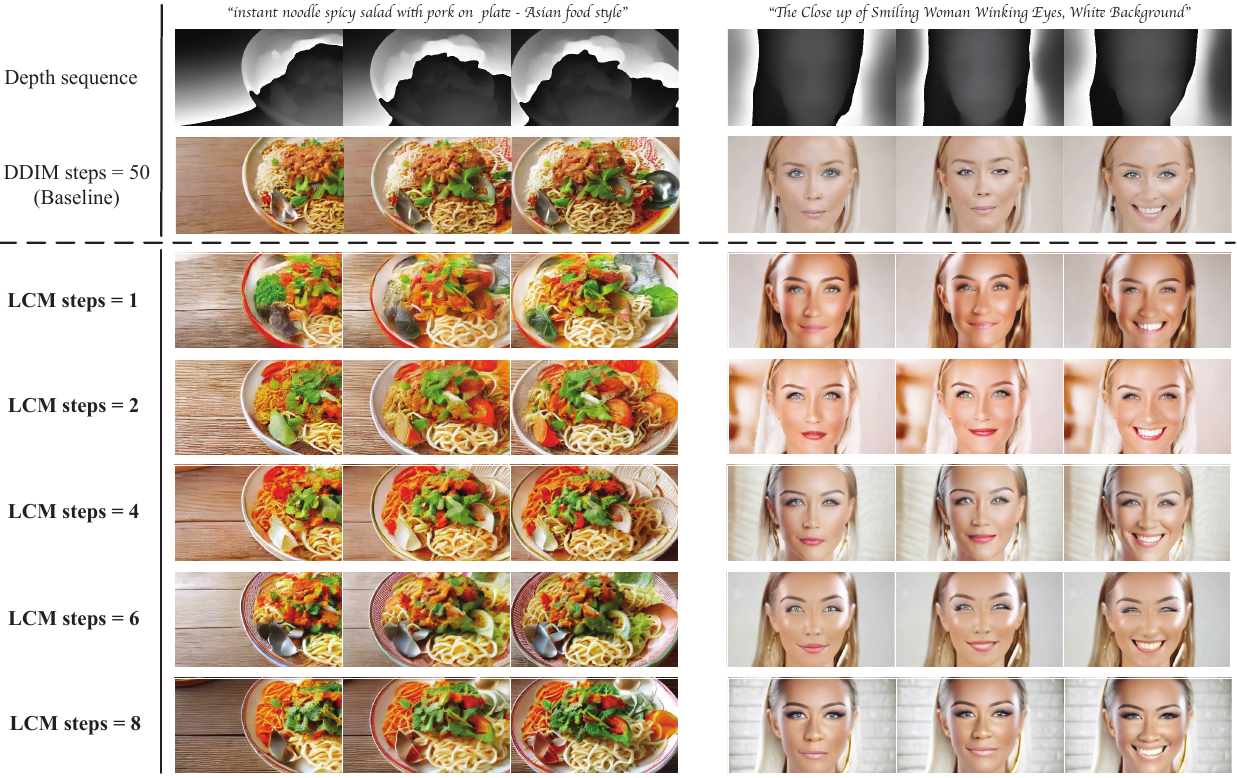}
  \vspace{-2mm}
    \caption{\textbf{Ablation study on compositional depth-to-video synthesis task under different inference steps}.
    Since the additional depth sequence can provide prior guidance about structure and temporal, our method can produce plausible results with fewer steps or even only one step.
    }
    \label{fig:ablation_steps_depth}
\end{figure*}

\begin{table}[t]
    \caption{
        {\textbf{Inference latency} on text-to-video generation task.
        All experiments are performed on an NVIDIA A100 GPU. 
        The inference overhead of generating eight videos at a time is reported.
    }}
    \label{tab:inference_cost}
    \vspace{-2mm}
    \tablestyle{5pt}{1.1}
    \centering
    \setlength{\tabcolsep}{6.0pt}{
      \begin{tabular}{l|ccc}
       { Method} & Step & Resolution & Latency \\
                                
        \shline
        Baseline & DDIM 50-step    & $16\times256\times256$   &  60s \\
         \rowcolor{Gray}
        \method & LCM 4-step  & $16\times256\times256$  & \textbf{10s} \\
        \shline
        Baseline & DDIM 50-step   & $16\times448\times256$   &  104s \\
         \rowcolor{Gray}
        \method & LCM 4-step  & $16\times448\times256$  & \textbf{16s} \\
 %
%
    \end{tabular}
    }
    \vspace{-2mm}
\end{table}

\begin{figure*}
  \centering
  \includegraphics[width=0.99\linewidth]{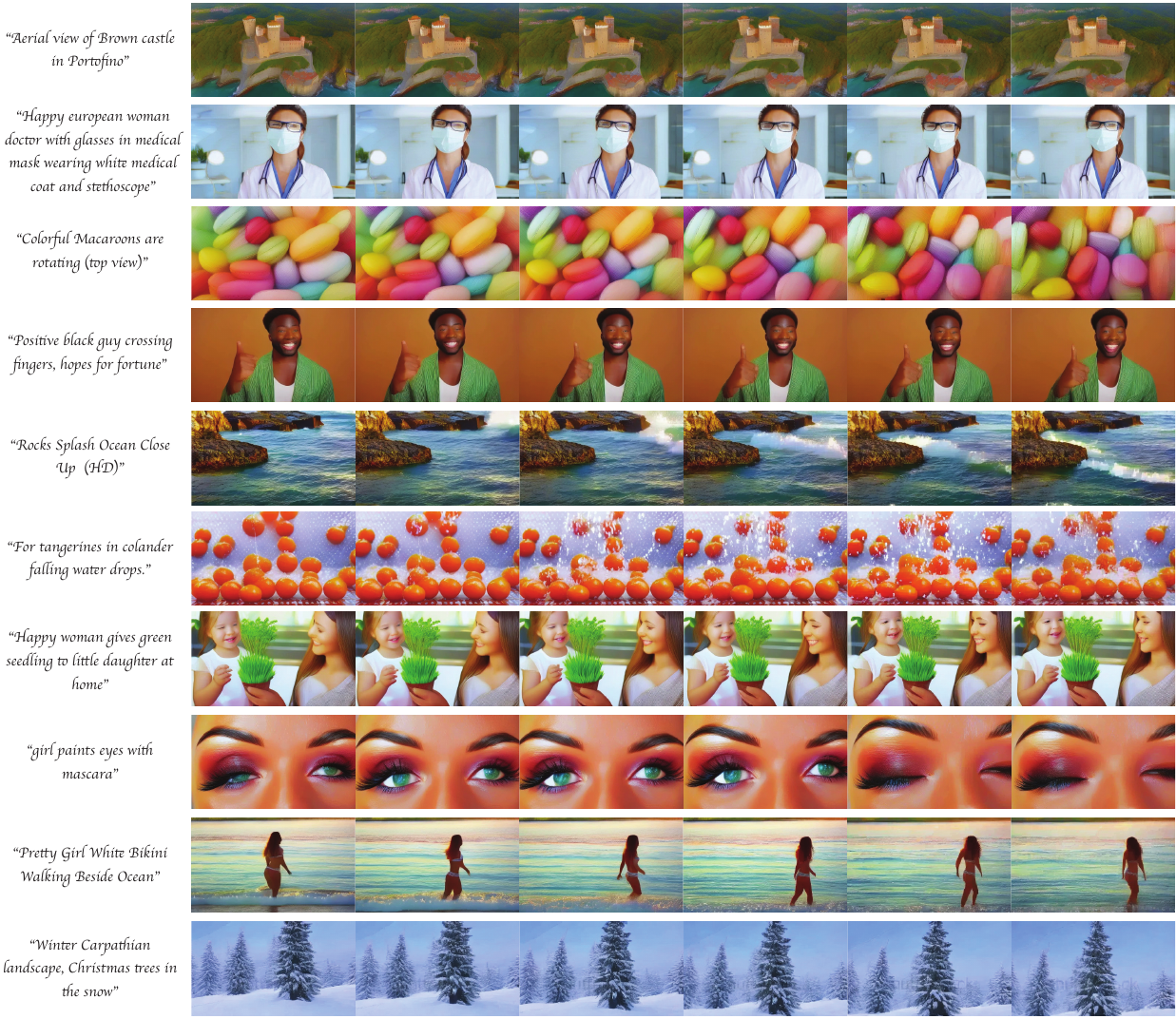}
  \vspace{-2mm}
    \caption{\textbf{Qualitative visualization results on text-to-video generation task}. Videos are synthesized by performing 4 denoising steps.}
    \label{fig:more_results_t2v}
\end{figure*}

\begin{figure*}
  \centering
  \includegraphics[width=0.99\linewidth]{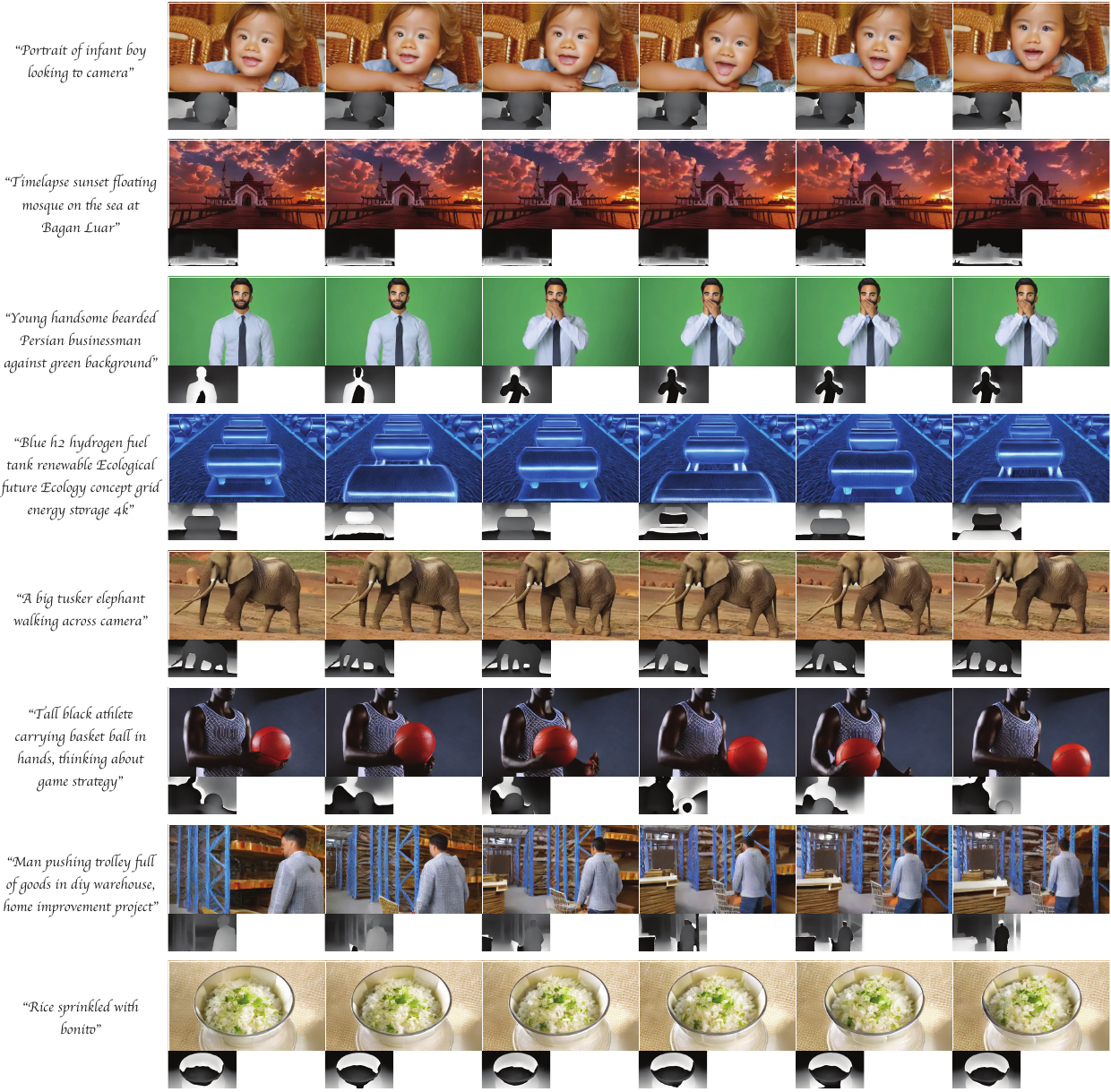}
  \vspace{-3mm}
    \caption{\textbf{Qualitative results on compositional depth-to-video synthesis task}. Videos are synthesized by performing 4 denoising steps.}
    \label{fig:more_results_depth}
\end{figure*}

\begin{figure*}
  \centering
  \includegraphics[width=0.99\linewidth]{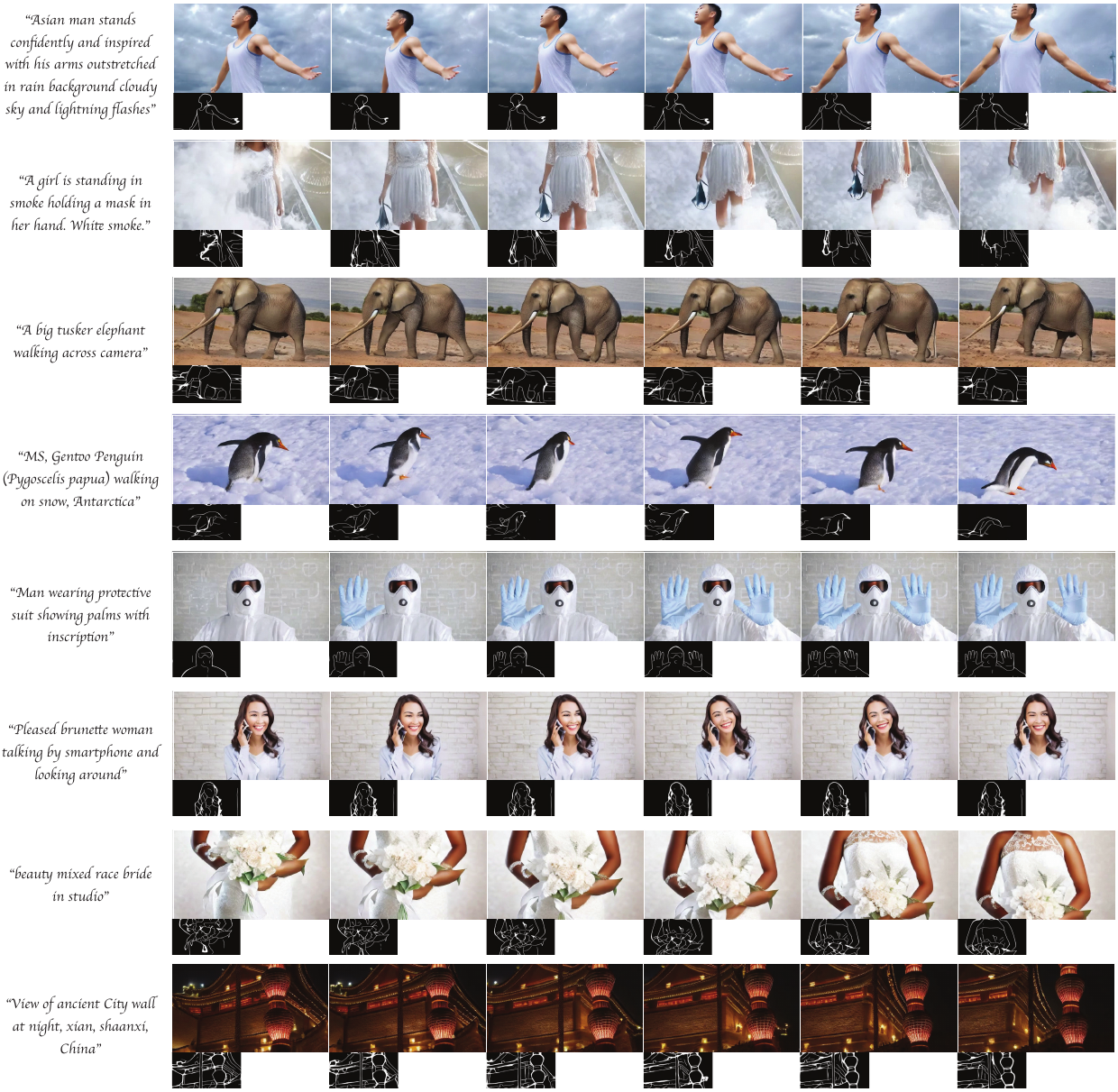}
  \vspace{-2mm}
    \caption{\textbf{Qualitative results on compositional sketch-to-video synthesis task}. Videos are synthesized by performing 4 denoising steps.}
    \label{fig:more_results_sketch}
\end{figure*}

\begin{figure*}
  \centering
  \includegraphics[width=0.99\linewidth]{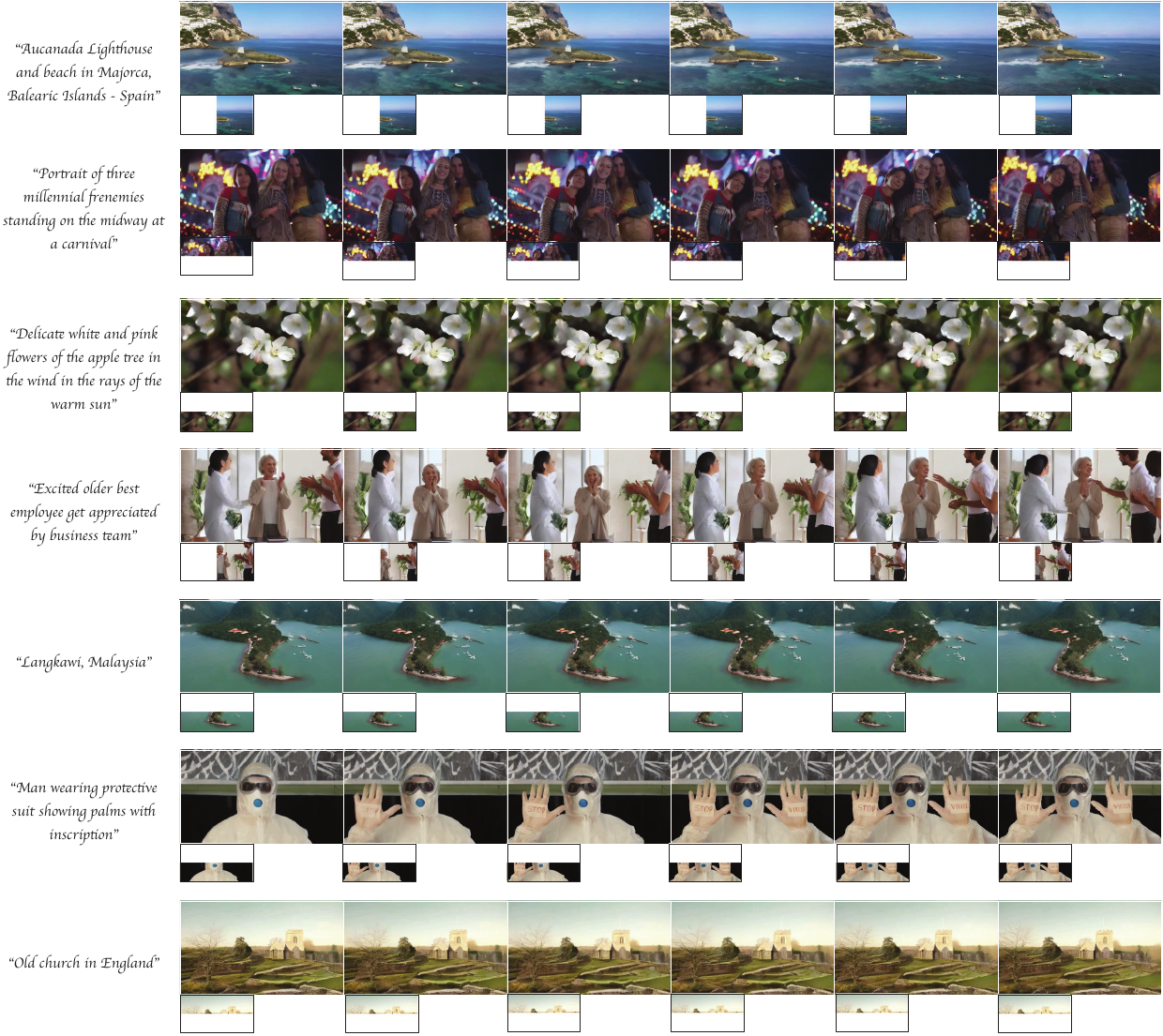}
  \vspace{-2mm}
    \caption{\textbf{Qualitative visualizations on compositional video inpainting task}. Videos are synthesized by performing 4 denoising steps.}
    \label{fig:more_results_inpainting}
\end{figure*}


\subsection{Time efficiency}

We measure the inference time required for text-to-video synthesis using our proposed \method and compare it with the baseline method~\cite{TFT2V}.
The comparative results are exhibited in \cref{tab:inference_cost}.
From the results, we can observe that since our method merely requires 4 LCM steps for inference, it is much faster than the baseline method with 50 DDIM steps. 
It should be noted that in addition to iterative denoising, the inference cost also includes text feature encoding, latent code decoding, \etc.
In addition, we notice that \method saves more time on high-resolution generation compared to baseline.
For example, generating a video of $16\times 256 \times 256 $ size saves 50 seconds (\ie, $60s-10s$), while 16x448x256 saves 88 seconds.
The above comparison demonstrates the high efficiency of our approach.


\subsection{Ablation study on inference steps}

In \cref{fig:ablation_steps_t2v}, we present the experimental visualization of varying inference steps in the text-to-video task.
The results indicate that when the sampling step is too small, such as step $=$ 1, the generated videos suffer from blurriness, with many details being inaccurately represented, and the temporal structure of objects cannot be preserved.
As the number of iteration steps increases, the visual quality gradually improves, and the temporal structure can be better maintained.
For example, when using 4$\sim$6 steps, comparable results to DDIM's 50 steps can be achieved, while significantly reducing the sampling steps and improving the generation speed.
Moreover, we additionally perform an ablation study on compositional depth-to-video synthesis. As illustrated in \cref{fig:ablation_steps_depth}, we observe that with only a few steps, such as 1 step, the generated results can display good visual quality. With an increase in the step size, many details became even more apparent. 
We attribute this phenomenon to the fact that compared to text-to-video generation, compositional depth-to-video synthesis can leverage additional structured control information, which reduces the difficulty of predicting videos from pure noise and enables achieving high-quality results with fewer steps. Empirically, we find that 2$\sim$4 steps can produce relatively stable compositional synthesis contents. 
The results on both text-to-video and compositional video generation demonstrate the effectiveness of the proposed \method, achieving a good balance between quality and speed.

\subsection{More qualitative results}

To comprehensively showcase the capabilities of our method in handling various video generation tasks, we present additional visualization examples in \cref{fig:more_results_t2v}. These results are generated with 4 LCM steps. 
From the visualization results, it is evident that our method achieves impressive generation outcomes in terms of visual quality and temporal continuity. 

In \cref{fig:more_results_depth,fig:more_results_sketch,fig:more_results_inpainting}, we present more results of compositional video generation using 4 sampling steps, including compositional depth-to-video synthesis (\cref{fig:more_results_depth}), compositional sketch-to-video synthesis (\cref{fig:more_results_sketch}), and compositional video inpainting (\cref{fig:more_results_inpainting}). 
These qualitative results demonstrate stable generation quality and highlight the controllability of our method in accordance with the input conditioning.
Our method can be applied to multiple mainstream video generation tasks mentioned above, revealing the generality of the \method framework and the vast potential for other future applications.

\subsection{Limitations}

In our \method, we explore the application of the consistency model in video generation tasks, including text-to-video generation and compositional video synthesis. 
However, there are certain limitations: 1) Our method relies on a strong teacher model as the distillation target. 2) The consistency distillation process requires finetuning the model. While consistency distillation only requires a small number of training steps, it may lead to unsatisfactory results when the training data for the teacher model is unavailable or from different domains. 3) Even though our method reduces the inference steps to 4$\sim$6, real-time video generation, like image generation, is still not achieved. 
Exploring more stable and efficient video generation algorithms while ensuring the high quality of the generated video content is a promising future direction. 

%% file: sec/5_conclusion.tex
\section{Conclusion}

In this work, we propose the \method framework that extents latent consistency models to the video generation field. 
Our approach leverages existing latent video diffusion models and employs the consistency distillation technique to enable efficient and fast video synthesis.
%
Experimental results demonstrate the effectiveness of our approach, with high-fidelity video synthesis achieved in just four steps, showcasing its real-time synthesis potential compared to prior methods requiring approximately 50 DDIM steps.
%
We hope that \method can serve as a simple yet
effective baseline for subsequent research work.

\vspace{2mm}
\noindent{\textbf{Acknowledgements.}}
This work is supported by the National Natural Science Foundation
of China under grant U22B2053 and Alibaba Group through Alibaba Research Intern Program.

%

%% file: main.bbl
\begin{thebibliography}{61}
\providecommand{\natexlab}[1]{#1}
\providecommand{\url}[1]{\texttt{#1}}
\expandafter\ifx\csname urlstyle\endcsname\relax
  \providecommand{\doi}[1]{doi: #1}\else
  \providecommand{\doi}{doi: \begingroup \urlstyle{rm}\Url}\fi

\bibitem[An et~al.(2023)An, Zhang, Yang, Gupta, Huang, Luo, and Yin]{an2023latent}
Jie An, Songyang Zhang, Harry Yang, Sonal Gupta, Jia-Bin Huang, Jiebo Luo, and Xi Yin.
\newblock Latent-shift: Latent diffusion with temporal shift for efficient text-to-video generation.
\newblock \emph{arXiv preprint arXiv:2304.08477}, 2023.

\bibitem[Bain et~al.(2021)Bain, Nagrani, Varol, and Zisserman]{webvid10m}
Max Bain, Arsha Nagrani, G{\"u}l Varol, and Andrew Zisserman.
\newblock Frozen in time: A joint video and image encoder for end-to-end retrieval.
\newblock In \emph{ICCV}, pages 1728--1738, 2021.

\bibitem[Balaji et~al.(2019)Balaji, Min, Bai, Chellappa, and Graf]{balaji2019conditional}
Yogesh Balaji, Martin~Renqiang Min, Bing Bai, Rama Chellappa, and Hans~Peter Graf.
\newblock Conditional {GAN} with discriminative filter generation for text-to-video synthesis.
\newblock In \emph{IJCAI}, page~2, 2019.

\bibitem[Blattmann et~al.(2023{\natexlab{a}})Blattmann, Dockhorn, Kulal, Mendelevitch, Kilian, Lorenz, Levi, English, Voleti, Letts, et~al.]{SVD}
Andreas Blattmann, Tim Dockhorn, Sumith Kulal, Daniel Mendelevitch, Maciej Kilian, Dominik Lorenz, Yam Levi, Zion English, Vikram Voleti, Adam Letts, et~al.
\newblock Stable video diffusion: Scaling latent video diffusion models to large datasets.
\newblock \emph{arXiv preprint arXiv:2311.15127}, 2023{\natexlab{a}}.

\bibitem[Blattmann et~al.(2023{\natexlab{b}})Blattmann, Rombach, Ling, Dockhorn, Kim, Fidler, and Kreis]{VideoLDM}
Andreas Blattmann, Robin Rombach, Huan Ling, Tim Dockhorn, Seung~Wook Kim, Sanja Fidler, and Karsten Kreis.
\newblock Align your latents: High-resolution video synthesis with latent diffusion models.
\newblock In \emph{CVPR}, pages 22563--22575, 2023{\natexlab{b}}.

\bibitem[Chai et~al.(2023)Chai, Guo, Wang, and Lu]{chai2023stablevideo}
Wenhao Chai, Xun Guo, Gaoang Wang, and Yan Lu.
\newblock Stablevideo: Text-driven consistency-aware diffusion video editing.
\newblock In \emph{ICCV}, pages 23040--23050, 2023.

\bibitem[Chang et~al.(2023)Chang, Zhang, Barber, Maschinot, Lezama, Jiang, Yang, Murphy, Freeman, Rubinstein, et~al.]{chang2023muse}
Huiwen Chang, Han Zhang, Jarred Barber, AJ Maschinot, Jose Lezama, Lu Jiang, Ming-Hsuan Yang, Kevin Murphy, William~T Freeman, Michael Rubinstein, et~al.
\newblock Muse: Text-to-image generation via masked generative {T}ransformers.
\newblock \emph{arXiv preprint arXiv:2301.00704}, 2023.

\bibitem[Chen et~al.(2023{\natexlab{a}})Chen, Lin, Tseng, Lin, and Yang]{chen2023motion}
Tsai-Shien Chen, Chieh~Hubert Lin, Hung-Yu Tseng, Tsung-Yi Lin, and Ming-Hsuan Yang.
\newblock Motion-conditioned diffusion model for controllable video synthesis.
\newblock \emph{arXiv preprint arXiv:2304.14404}, 2023{\natexlab{a}}.

\bibitem[Chen et~al.(2023{\natexlab{b}})Chen, Wu, Xie, Wu, Li, Xia, Xiao, and Lin]{chen2023control}
Weifeng Chen, Jie Wu, Pan Xie, Hefeng Wu, Jiashi Li, Xin Xia, Xuefeng Xiao, and Liang Lin.
\newblock Control-a-video: Controllable text-to-video generation with diffusion models.
\newblock \emph{arXiv preprint arXiv:2305.13840}, 2023{\natexlab{b}}.

\bibitem[Chen et~al.(2023{\natexlab{c}})Chen, Huang, Liu, Shen, Zhao, and Zhao]{chen2023anydoor}
Xi Chen, Lianghua Huang, Yu Liu, Yujun Shen, Deli Zhao, and Hengshuang Zhao.
\newblock Anydoor: Zero-shot object-level image customization.
\newblock \emph{arXiv preprint arXiv:2307.09481}, 2023{\natexlab{c}}.

\bibitem[Esser et~al.(2023)Esser, Chiu, Atighehchian, Granskog, and Germanidis]{Gen-1}
Patrick Esser, Johnathan Chiu, Parmida Atighehchian, Jonathan Granskog, and Anastasis Germanidis.
\newblock Structure and content-guided video synthesis with diffusion models.
\newblock In \emph{ICCV}, pages 7346--7356, 2023.

\bibitem[Ge et~al.(2023)Ge, Nah, Liu, Poon, Tao, Catanzaro, Jacobs, Huang, Liu, and Balaji]{ge2023preserve}
Songwei Ge, Seungjun Nah, Guilin Liu, Tyler Poon, Andrew Tao, Bryan Catanzaro, David Jacobs, Jia-Bin Huang, Ming-Yu Liu, and Yogesh Balaji.
\newblock Preserve your own correlation: A noise prior for video diffusion models.
\newblock In \emph{ICCV}, pages 22930--22941, 2023.

\bibitem[Goodfellow et~al.(2014)Goodfellow, Pouget-Abadie, Mirza, Xu, Warde-Farley, Ozair, Courville, and Bengio]{goodfellow2014generative}
Ian Goodfellow, Jean Pouget-Abadie, Mehdi Mirza, Bing Xu, David Warde-Farley, Sherjil Ozair, Aaron Courville, and Yoshua Bengio.
\newblock Generative adversarial nets.
\newblock \emph{NeurIPS}, 27, 2014.

\bibitem[Guo et~al.(2023)Guo, Yang, Rao, Wang, Qiao, Lin, and Dai]{guo2023animatediff}
Yuwei Guo, Ceyuan Yang, Anyi Rao, Yaohui Wang, Yu Qiao, Dahua Lin, and Bo Dai.
\newblock Animatediff: Animate your personalized text-to-image diffusion models without specific tuning.
\newblock \emph{arXiv preprint arXiv:2307.04725}, 2023.

\bibitem[Ho and Salimans(2022)]{ho2022classifierfreeguidance}
Jonathan Ho and Tim Salimans.
\newblock Classifier-free diffusion guidance.
\newblock \emph{arXiv preprint arXiv:2207.12598}, 2022.

\bibitem[Ho et~al.(2022)Ho, Chan, Saharia, Whang, Gao, Gritsenko, Kingma, Poole, Norouzi, Fleet, et~al.]{imagenvideo}
Jonathan Ho, William Chan, Chitwan Saharia, Jay Whang, Ruiqi Gao, Alexey Gritsenko, Diederik~P Kingma, Ben Poole, Mohammad Norouzi, David~J Fleet, et~al.
\newblock Imagen video: High definition video generation with diffusion models.
\newblock \emph{arXiv preprint arXiv:2210.02303}, 2022.

\bibitem[Hu et~al.(2023)Hu, Gao, Zhang, Sun, Zhang, and Bo]{hu2023animate}
Li Hu, Xin Gao, Peng Zhang, Ke Sun, Bang Zhang, and Liefeng Bo.
\newblock Animate anyone: Consistent and controllable image-to-video synthesis for character animation.
\newblock \emph{arXiv preprint arXiv:2311.17117}, 2023.

\bibitem[Huang et~al.(2023)Huang, Chen, Liu, Shen, Zhao, and Zhou]{huang2023composer}
Lianghua Huang, Di Chen, Yu Liu, Yujun Shen, Deli Zhao, and Jingren Zhou.
\newblock Composer: Creative and controllable image synthesis with composable conditions.
\newblock \emph{ICML}, 2023.

\bibitem[Kang et~al.(2023)Kang, Zhu, Zhang, Park, Shechtman, Paris, and Park]{kang2023scaling}
Minguk Kang, Jun-Yan Zhu, Richard Zhang, Jaesik Park, Eli Shechtman, Sylvain Paris, and Taesung Park.
\newblock Scaling up {GANs} for text-to-image synthesis.
\newblock In \emph{CVPR}, pages 10124--10134, 2023.

\bibitem[Kawar et~al.(2023)Kawar, Zada, Lang, Tov, Chang, Dekel, Mosseri, and Irani]{kawar2023imagic}
Bahjat Kawar, Shiran Zada, Oran Lang, Omer Tov, Huiwen Chang, Tali Dekel, Inbar Mosseri, and Michal Irani.
\newblock Imagic: Text-based real image editing with diffusion models.
\newblock In \emph{CVPR}, pages 6007--6017, 2023.

\bibitem[Khachatryan et~al.(2023)Khachatryan, Movsisyan, Tadevosyan, Henschel, Wang, Navasardyan, and Shi]{Text2video-zero}
Levon Khachatryan, Andranik Movsisyan, Vahram Tadevosyan, Roberto Henschel, Zhangyang Wang, Shant Navasardyan, and Humphrey Shi.
\newblock Text2video-zero: Text-to-image diffusion models are zero-shot video generators.
\newblock \emph{arXiv preprint arXiv:2303.13439}, 2023.

\bibitem[Liu et~al.(2023)Liu, Feng, Zhu, Zhang, Zheng, Liu, Zhao, Zhou, and Cao]{liu2023cones}
Zhiheng Liu, Ruili Feng, Kai Zhu, Yifei Zhang, Kecheng Zheng, Yu Liu, Deli Zhao, Jingren Zhou, and Yang Cao.
\newblock Cones: Concept neurons in diffusion models for customized generation.
\newblock In \emph{ICML}, 2023.

\bibitem[Luo et~al.(2023{\natexlab{a}})Luo, Tan, Huang, Li, and Zhao]{LCM}
Simian Luo, Yiqin Tan, Longbo Huang, Jian Li, and Hang Zhao.
\newblock Latent consistency models: Synthesizing high-resolution images with few-step inference.
\newblock \emph{arXiv preprint arXiv:2310.04378}, 2023{\natexlab{a}}.

\bibitem[Luo et~al.(2023{\natexlab{b}})Luo, Tan, Patil, Gu, von Platen, Passos, Huang, Li, and Zhao]{LCM_lora}
Simian Luo, Yiqin Tan, Suraj Patil, Daniel Gu, Patrick von Platen, Apolin{\'a}rio Passos, Longbo Huang, Jian Li, and Hang Zhao.
\newblock Lcm-lora: A universal stable-diffusion acceleration module.
\newblock \emph{arXiv preprint arXiv:2311.05556}, 2023{\natexlab{b}}.

\bibitem[Luo et~al.(2023{\natexlab{c}})Luo, Chen, Zhang, Huang, Wang, Shen, Zhao, Zhou, and Tan]{luo2023videofusion}
Zhengxiong Luo, Dayou Chen, Yingya Zhang, Yan Huang, Liang Wang, Yujun Shen, Deli Zhao, Jingren Zhou, and Tieniu Tan.
\newblock Videofusion: Decomposed diffusion models for high-quality video generation.
\newblock In \emph{CVPR}, pages 10209--10218, 2023{\natexlab{c}}.

\bibitem[Molad et~al.(2023)Molad, Horwitz, Valevski, Acha, Matias, Pritch, Leviathan, and Hoshen]{molad2023dreamix}
Eyal Molad, Eliahu Horwitz, Dani Valevski, Alex~Rav Acha, Yossi Matias, Yael Pritch, Yaniv Leviathan, and Yedid Hoshen.
\newblock Dreamix: Video diffusion models are general video editors.
\newblock \emph{arXiv preprint arXiv:2302.01329}, 2023.

\bibitem[Mou et~al.(2023)Mou, Wang, Xie, Zhang, Qi, Shan, and Qie]{T2i-adapter}
Chong Mou, Xintao Wang, Liangbin Xie, Jian Zhang, Zhongang Qi, Ying Shan, and Xiaohu Qie.
\newblock T2i-adapter: Learning adapters to dig out more controllable ability for text-to-image diffusion models.
\newblock \emph{arXiv preprint arXiv:2302.08453}, 2023.

\bibitem[Ni et~al.(2023)Ni, Shi, Li, Huang, and Min]{ni2023conditional}
Haomiao Ni, Changhao Shi, Kai Li, Sharon~X Huang, and Martin~Renqiang Min.
\newblock Conditional image-to-video generation with latent flow diffusion models.
\newblock In \emph{CVPR}, pages 18444--18455, 2023.

\bibitem[Nichol et~al.(2022)Nichol, Dhariwal, Ramesh, Shyam, Mishkin, Mcgrew, Sutskever, and Chen]{GLIDE}
Alexander~Quinn Nichol, Prafulla Dhariwal, Aditya Ramesh, Pranav Shyam, Pamela Mishkin, Bob Mcgrew, Ilya Sutskever, and Mark Chen.
\newblock Glide: Towards photorealistic image generation and editing with text-guided diffusion models.
\newblock In \emph{ICML}, pages 16784--16804. PMLR, 2022.

\bibitem[Ramesh et~al.(2022)Ramesh, Dhariwal, Nichol, Chu, and Chen]{Dalle2}
Aditya Ramesh, Prafulla Dhariwal, Alex Nichol, Casey Chu, and Mark Chen.
\newblock Hierarchical text-conditional image generation with clip latents.
\newblock \emph{arXiv preprint arXiv:2204.06125}, 1\penalty0 (2):\penalty0 3, 2022.

\bibitem[Rombach et~al.(2022)Rombach, Blattmann, Lorenz, Esser, and Ommer]{stablediffusion}
Robin Rombach, Andreas Blattmann, Dominik Lorenz, Patrick Esser, and Bj{\"o}rn Ommer.
\newblock High-resolution image synthesis with latent diffusion models.
\newblock In \emph{CVPR}, pages 10684--10695, 2022.

\bibitem[Ruiz et~al.(2023)Ruiz, Li, Jampani, Pritch, Rubinstein, and Aberman]{ruiz2023dreambooth}
Nataniel Ruiz, Yuanzhen Li, Varun Jampani, Yael Pritch, Michael Rubinstein, and Kfir Aberman.
\newblock Dreambooth: Fine tuning text-to-image diffusion models for subject-driven generation.
\newblock In \emph{CVPR}, pages 22500--22510, 2023.

\bibitem[Saharia et~al.(2022)Saharia, Chan, Saxena, Li, Whang, Denton, Ghasemipour, Gontijo~Lopes, Karagol~Ayan, Salimans, et~al.]{saharia2022photorealistic}
Chitwan Saharia, William Chan, Saurabh Saxena, Lala Li, Jay Whang, Emily~L Denton, Kamyar Ghasemipour, Raphael Gontijo~Lopes, Burcu Karagol~Ayan, Tim Salimans, et~al.
\newblock Photorealistic text-to-image diffusion models with deep language understanding.
\newblock \emph{NeurIPS}, 35:\penalty0 36479--36494, 2022.

\bibitem[Salimans and Ho(2022)]{vprediction}
Tim Salimans and Jonathan Ho.
\newblock Progressive distillation for fast sampling of diffusion models.
\newblock \emph{arXiv preprint arXiv:2202.00512}, 2022.

\bibitem[Sauer et~al.(2023)Sauer, Lorenz, Blattmann, and Rombach]{sauer2023adversarial}
Axel Sauer, Dominik Lorenz, Andreas Blattmann, and Robin Rombach.
\newblock Adversarial diffusion distillation.
\newblock \emph{arXiv preprint arXiv:2311.17042}, 2023.

\bibitem[Schuhmann et~al.(2022)Schuhmann, Beaumont, Vencu, Gordon, Wightman, Cherti, Coombes, Katta, Mullis, Wortsman, et~al.]{schuhmann2022laion}
Christoph Schuhmann, Romain Beaumont, Richard Vencu, Cade Gordon, Ross Wightman, Mehdi Cherti, Theo Coombes, Aarush Katta, Clayton Mullis, Mitchell Wortsman, et~al.
\newblock Laion-5b: An open large-scale dataset for training next generation image-text models.
\newblock \emph{NeurIPS}, 35:\penalty0 25278--25294, 2022.

\bibitem[Shen and Zhou(2021)]{shen2021closed}
Yujun Shen and Bolei Zhou.
\newblock Closed-form factorization of latent semantics in {GANs}.
\newblock In \emph{CVPR}, pages 1532--1540, 2021.

\bibitem[Singer et~al.(2023)Singer, Polyak, Hayes, Yin, An, Zhang, Hu, Yang, Ashual, Gafni, et~al.]{make-a-video}
Uriel Singer, Adam Polyak, Thomas Hayes, Xi Yin, Jie An, Songyang Zhang, Qiyuan Hu, Harry Yang, Oron Ashual, Oran Gafni, et~al.
\newblock Make-a-video: Text-to-video generation without text-video data.
\newblock \emph{ICLR}, 2023.

\bibitem[Skorokhodov et~al.(2022)Skorokhodov, Tulyakov, and Elhoseiny]{skorokhodov2022stylegan}
Ivan Skorokhodov, Sergey Tulyakov, and Mohamed Elhoseiny.
\newblock Style{GAN}-v: A continuous video generator with the price, image quality and perks of {StyleGAN2}.
\newblock In \emph{CVPR}, pages 3626--3636, 2022.

\bibitem[Song et~al.(2021)Song, Meng, and Ermon]{DDIM}
Jiaming Song, Chenlin Meng, and Stefano Ermon.
\newblock Denoising diffusion implicit models.
\newblock In \emph{ICLR}, 2021.

\bibitem[Song et~al.(2023)Song, Dhariwal, Chen, and Sutskever]{consistencymodel}
Yang Song, Prafulla Dhariwal, Mark Chen, and Ilya Sutskever.
\newblock Consistency models.
\newblock In \emph{ICML}, 2023.

\bibitem[Tulyakov et~al.(2018)Tulyakov, Liu, Yang, and Kautz]{mocogan}
Sergey Tulyakov, Ming-Yu Liu, Xiaodong Yang, and Jan Kautz.
\newblock Moco{GAN}: Decomposing motion and content for video generation.
\newblock In \emph{CVPR}, pages 1526--1535, 2018.

\bibitem[Villegas et~al.(2022)Villegas, Babaeizadeh, Kindermans, Moraldo, Zhang, Saffar, Castro, Kunze, and Erhan]{villegas2022phenaki}
Ruben Villegas, Mohammad Babaeizadeh, Pieter-Jan Kindermans, Hernan Moraldo, Han Zhang, Mohammad~Taghi Saffar, Santiago Castro, Julius Kunze, and Dumitru Erhan.
\newblock Phenaki: Variable length video generation from open domain textual description.
\newblock \emph{arXiv preprint arXiv:2210.02399}, 2022.

\bibitem[Wang et~al.(2023{\natexlab{a}})Wang, Yuan, Chen, Zhang, Wang, and Zhang]{modelscopet2v}
Jiuniu Wang, Hangjie Yuan, Dayou Chen, Yingya Zhang, Xiang Wang, and Shiwei Zhang.
\newblock Modelscope text-to-video technical report.
\newblock \emph{arXiv preprint arXiv:2308.06571}, 2023{\natexlab{a}}.

\bibitem[Wang et~al.(2023{\natexlab{b}})Wang, Yang, Tuo, He, Zhu, Fu, and Liu]{wang2023videofactory}
Wenjing Wang, Huan Yang, Zixi Tuo, Huiguo He, Junchen Zhu, Jianlong Fu, and Jiaying Liu.
\newblock Videofactory: Swap attention in spatiotemporal diffusions for text-to-video generation.
\newblock \emph{arXiv preprint arXiv:2305.10874}, 2023{\natexlab{b}}.

\bibitem[Wang et~al.(2023{\natexlab{c}})Wang, Yuan, Zhang, Chen, Wang, Zhang, Shen, Zhao, and Zhou]{videocomposer}
Xiang Wang, Hangjie Yuan, Shiwei Zhang, Dayou Chen, Jiuniu Wang, Yingya Zhang, Yujun Shen, Deli Zhao, and Jingren Zhou.
\newblock Videocomposer: Compositional video synthesis with motion controllability.
\newblock \emph{NeurIPS}, 2023{\natexlab{c}}.

\bibitem[Wang et~al.(2023{\natexlab{d}})Wang, Zhang, Yuan, Qing, Gong, Zhang, Shen, Gao, and Sang]{TFT2V}
Xiang Wang, Shiwei Zhang, Hangjie Yuan, Zhiwu Qing, Biao Gong, Yingya Zhang, Yujun Shen, Changxin Gao, and Nong Sang.
\newblock A recipe for scaling up text-to-video generation with text-free videos.
\newblock \emph{arXiv}, 2023{\natexlab{d}}.

\bibitem[Wang et~al.(2020{\natexlab{a}})Wang, Bilinski, Bremond, and Dantcheva]{wang2020g3an}
Yaohui Wang, Piotr Bilinski, Francois Bremond, and Antitza Dantcheva.
\newblock G3an: Disentangling appearance and motion for video generation.
\newblock In \emph{CVPR}, pages 5264--5273, 2020{\natexlab{a}}.

\bibitem[Wang et~al.(2020{\natexlab{b}})Wang, Bilinski, Bremond, and Dantcheva]{wang2020imaginator}
Yaohui Wang, Piotr Bilinski, Francois Bremond, and Antitza Dantcheva.
\newblock Imaginator: Conditional spatio-temporal {GAN} for video generation.
\newblock In \emph{WACV}, pages 1160--1169, 2020{\natexlab{b}}.

\bibitem[Wang et~al.(2023{\natexlab{e}})Wang, Jiang, and Loy]{wang2023styleinv}
Yuhan Wang, Liming Jiang, and Chen~Change Loy.
\newblock Styleinv: A temporal style modulated inversion network for unconditional video generation.
\newblock In \emph{ICCV}, pages 22851--22861, 2023{\natexlab{e}}.

\bibitem[Wei et~al.(2023)Wei, Zhang, Qing, Yuan, Liu, Liu, Zhang, Zhou, and Shan]{wei2023dreamvideo}
Yujie Wei, Shiwei Zhang, Zhiwu Qing, Hangjie Yuan, Zhiheng Liu, Yu Liu, Yingya Zhang, Jingren Zhou, and Hongming Shan.
\newblock Dreamvideo: Composing your dream videos with customized subject and motion.
\newblock \emph{arXiv preprint arXiv:2312.04433}, 2023.

\bibitem[Wu et~al.(2023)Wu, Ge, Wang, Lei, Gu, Shi, Hsu, Shan, Qie, and Shou]{tune-a-video}
Jay~Zhangjie Wu, Yixiao Ge, Xintao Wang, Stan~Weixian Lei, Yuchao Gu, Yufei Shi, Wynne Hsu, Ying Shan, Xiaohu Qie, and Mike~Zheng Shou.
\newblock Tune-a-video: One-shot tuning of image diffusion models for text-to-video generation.
\newblock In \emph{ICCV}, pages 7623--7633, 2023.

\bibitem[Xiao et~al.(2023)Xiao, Zhu, Zhang, Liu, Shen, Liu, Fu, and Zha]{xiao2023ccm}
Jie Xiao, Kai Zhu, Han Zhang, Zhiheng Liu, Yujun Shen, Yu Liu, Xueyang Fu, and Zheng-Jun Zha.
\newblock Ccm: Adding conditional controls to text-to-image consistency models.
\newblock 2023.

\bibitem[Yin et~al.(2023)Yin, Wu, Liang, Shi, Li, Ming, and Duan]{yin2023dragnuwa}
Shengming Yin, Chenfei Wu, Jian Liang, Jie Shi, Houqiang Li, Gong Ming, and Nan Duan.
\newblock Dragnuwa: Fine-grained control in video generation by integrating text, image, and trajectory.
\newblock \emph{arXiv preprint arXiv:2308.08089}, 2023.

\bibitem[Yu et~al.(2023)Yu, Sohn, Kim, and Shin]{yu2023video}
Sihyun Yu, Kihyuk Sohn, Subin Kim, and Jinwoo Shin.
\newblock Video probabilistic diffusion models in projected latent space.
\newblock In \emph{CVPR}, pages 18456--18466, 2023.

\bibitem[Zhang et~al.(2023{\natexlab{a}})Zhang, Wu, Liu, Zhao, Ran, Gu, Gao, and Shou]{zhang2023show}
David~Junhao Zhang, Jay~Zhangjie Wu, Jia-Wei Liu, Rui Zhao, Lingmin Ran, Yuchao Gu, Difei Gao, and Mike~Zheng Shou.
\newblock Show-1: Marrying pixel and latent diffusion models for text-to-video generation.
\newblock \emph{arXiv preprint arXiv:2309.15818}, 2023{\natexlab{a}}.

\bibitem[Zhang et~al.(2023{\natexlab{b}})Zhang, Rao, and Agrawala]{controlnet}
Lvmin Zhang, Anyi Rao, and Maneesh Agrawala.
\newblock Adding conditional control to text-to-image diffusion models.
\newblock In \emph{ICCV}, pages 3836--3847, 2023{\natexlab{b}}.

\bibitem[Zhang et~al.(2023{\natexlab{c}})Zhang, Wang, Zhang, Zhao, Yuan, Qin, Wang, Zhao, and Zhou]{zhang2023i2vgen}
Shiwei Zhang, Jiayu Wang, Yingya Zhang, Kang Zhao, Hangjie Yuan, Zhiwu Qin, Xiang Wang, Deli Zhao, and Jingren Zhou.
\newblock I2vgen-xl: High-quality image-to-video synthesis via cascaded diffusion models.
\newblock \emph{arXiv preprint arXiv:2311.04145}, 2023{\natexlab{c}}.

\bibitem[Zhang et~al.(2023{\natexlab{d}})Zhang, Wei, Jiang, Zhang, Zuo, and Tian]{zhang2023controlvideo}
Yabo Zhang, Yuxiang Wei, Dongsheng Jiang, Xiaopeng Zhang, Wangmeng Zuo, and Qi Tian.
\newblock Controlvideo: Training-free controllable text-to-video generation.
\newblock \emph{arXiv preprint arXiv:2305.13077}, 2023{\natexlab{d}}.

\bibitem[Zhao et~al.(2023)Zhao, Wang, Bao, Li, and Zhu]{zhao2023controlvideo}
Min Zhao, Rongzhen Wang, Fan Bao, Chongxuan Li, and Jun Zhu.
\newblock Controlvideo: Adding conditional control for one shot text-to-video editing.
\newblock \emph{arXiv preprint arXiv:2305.17098}, 2023.

\bibitem[Zhou et~al.(2022)Zhou, Wang, Yan, Lv, Zhu, and Feng]{zhou2022magicvideo}
Daquan Zhou, Weimin Wang, Hanshu Yan, Weiwei Lv, Yizhe Zhu, and Jiashi Feng.
\newblock Magicvideo: Efficient video generation with latent diffusion models.
\newblock \emph{arXiv preprint arXiv:2211.11018}, 2022.

\end{thebibliography}
